\documentclass[sii]{ipart}

\usepackage{natbib}
\usepackage{amsmath,amsfonts,amsthm,amscd,amssymb,bm,url,color,latexsym}

\allowdisplaybreaks

\renewcommand{\mathbf}{\boldsymbol}

\newcommand{\mb}{\mathbf}

\newcommand{ \paren }[1]{\left( #1 \right)}

\numberwithin{equation}{section}

\usepackage{graphicx}
\usepackage{url}
\usepackage{pifont}
\newcommand{\cmark}{\ding{51}}%
\newcommand{\xmark}{\ding{55}}%
\newcommand{\done}{\rlap{$\square$}{\raisebox{2pt}{\large\hspace{1pt}\cmark}}%
\hspace{-2.5pt}}
\newcommand{\tobedone}{\rlap{$\square$}{\large\hspace{1pt}\xmark}}
\usepackage{algorithm}
\usepackage{algpseudocode}
\usepackage{tablefootnote}
\usepackage{multirow}
\usepackage[skip=0pt]{caption}
\usepackage[hidelinks]{hyperref}
\usepackage[capitalize,nameinlink]{cleveref}

\startlocaldefs
\endlocaldefs

\pubyear{2022}
\volume{0}
\issue{0}
\firstpage{1}
\lastpage{1}

\begin{document}

\begin{frontmatter}

\title{Welfare and Fairness Dynamics in Federated Learning: A Client Selection Perspective}


\begin{aug}

\author{\inits{Y.}\fnms{Yash} \snm{Travadi}\ead[label=e1]{trava029@umn.edu}},
    \address{School of Statistics\\
    		University of Minnesota\\
            313 Ford Hall\\
            224 Church St SE\\
            Minneapolis, MN 55455, United States\\
             \printead{e1}}
    \author{\inits{L.}\fnms{Le} \snm{Peng}\ead[label=e2]{peng0347@umn.edu}},
    \address{Department of Computer Science \& Engineering\\
    		University of Minnesota\\
    	4-192 Keller Hall\\
            200 Union Street SE\\
            Minneapolis, MN 55455, United States\\
             \printead{e2}}
    \author{\inits{X.}\fnms{Xuan} \snm{Bi}\thanksref{t2}
            \ead[label=e3]{xbi@umn.edu}},
    \address{ Information \& Decision Sciences Department\\
            Carlson School of Management\\
    		University of Minnesota\\
            321 Nineteenth Avenue South\\
            Minneapolis, MN 55455, United States\\
             \printead{e3}}
    \author{\inits{J.}\fnms{Ju} \snm{Sun}\ead[label=e4]{jusun@umn.edu}}
    \address{Department of Computer Science \& Engineering\\
    	University of Minnesota\\
    	4-192 Keller Hall\\
            200 Union Street SE\\
            Minneapolis, MN 55455, United States\\
             \printead{e4}}
    \and
    \author{\inits{M.}\fnms{Mochen} \snm{Yang}\ead[label=e5]{yang3653@umn.edu}}
    \address{ Information \& Decision Sciences Department\\
            Carlson School of Management\\
    		University of Minnesota\\
            321 Nineteenth Avenue South\\
            Minneapolis, MN 55455, United States\\
             \printead{e5}}
    \thankstext{t2}{Corresponding author.}
\end{aug}
\received{\sday{26} \smonth{11} \syear{2022}}

\begin{abstract}
Federated learning (FL) is a privacy-preserving learning technique that enables distributed computing devices to train shared learning models across data silos collaboratively. Existing FL works mostly focus on designing advanced FL algorithms to improve the model performance. However, the economic considerations of the clients, such as fairness and incentive, are yet to be fully explored. Without such considerations, self-motivated clients may lose interest and leave the federation. To address this problem, we designed a novel incentive mechanism that involves a client selection process to remove low-quality clients and a money transfer process to ensure a fair reward distribution. Our experimental results strongly demonstrate that the proposed incentive mechanism can effectively improve the duration and fairness of the federation. 
\end{abstract}




\end{frontmatter}

\section{Introduction}
Artificial intelligence (AI) has revolutionized many aspects of our lives, such as transportation~\citep{ma2015large}, finance~\citep{ding2015deep}, healthcare~\citep{rajkomar2018scalable}, and more~\citep{nguyen2020monitoring, yunus2018framework, gopalakrishnan2018crack}. However, training generalizable AI models, especially deep neural networks (DNNs), typically requires massive amounts of data---the need escalates as the models get deeper and larger. In domains such as healthcare and finance, each site may have only limited data, and training large AI models requires data aggregation. However, data regulation and privacy concerns prohibit direct data sharing. Given these challenges, federated learning (FL)~\citep{konevcny2016federated}, which performs collaborative model training without direct data sharing, has emerged as a promising privacy-respecting learning paradigm. In FL, data remain in participant data shelters, and intermediate models trained locally are frequently aggregated and redistributed to facilitate information exchange. 

FL is a distributed learning paradigm that enables decentralized edge devices to collaboratively train machine learning models without data sharing while at a negligible loss of performance compared to centralized training. 

The most common FL framework represents a hub-and-spoke topology with a global server navigating the model training at each client by exchanging the local updates, such as gradients or model weights. Notably, throughout the training process, there is no raw data exchange, which greatly minimizes the risk of data leakage.

Despite the great success that FL has achieved~\citep{hard2018federated, li2019privacy, wang2019edge, li2019abnormal}, recent works identify some open problems in FL, including incentive~\citep{kairouz2021advances}. Incentive problems in FL may be caused by several factors. For example, clients need to invest computation and communication resources to support the federation, while the rewards may not be realized immediately.  The delay of reward can potentially induce FL to deteriorate before it reaches the desired goal. In addition, different clients may provide data with varied quality. If the reward is distributed unfairly, clients with high data quality may loss the incentive to contribute. Therefore, in principle, FL needs to be fair to ensure the benefits clients receive are commensurate with their data contributions and investments. Moreover, some clients may act as free riders in the federation, e.g., they may contribute only a small portion of their data or do not run enough local updates on their data~\citep{lin2019free}. These free-riding clients who still want to benefit from participating in the federation should be removed from the federation by some client selection mechanism. All these issues call for designing an incentivized FL scheme with fairness considerations. 

To promote a long-lasting federation and attract high-quality data contributors to remain in the loop, we propose an incentivized FL framework where high-quality clients are motivated to continuously contribute to the system, whereas low-quality clients are removed. Specifically, we quantify clients' fair reward distribution by measuring their contribution to the model performance under a budget balance constraint. By filtering low-quality clients through a client selection process, we show that our method can maximize social welfare (total gain of the federation) and promote a longer-term partnership.

Our main contribution includes: we \textbf{(1)} design a client selection scheme to remove the low-contributing clients which leads to higher social welfare, \textbf{(2)} propose a money transfer scheme to redistribute the reward to improve the fairness, and \textbf{(3)} conduct extensive simulations on both homogeneous and heterogeneous settings. The results consistently show the superiority of our proposed method. 

\section{Related Work}
In this section, we will provide a brief review of recent studies on FL and related works on incentive mechanism design.

\subsection{Federated learning}
The concept of FL was first introduced by Google~\citep{konevcny2016federated} in 2016 and applied to the design of a virtual keyboard application named Gboard~\citep{hard2018federated}. In an FL process, decentralized participants can jointly train a model without data sharing. This privacy-preserving property can potentially revolutionize many critical domains such as medicine~\citep{dayan2021federated, roth2020federated}, and finance~\citep{li2019abnormal, long2020federated}. The existing application of FL can be categorized into cross-silo or cross-device FL based on the scale of participants. In a cross-silo setting, the model is trained on a few distributed data centers with siloed data, commonly seen in medical and financial applications. In a cross-device setting, in contrast, the model is trained by a lot of edge devices such as mobile phones, where scalability is a major concern~\citep{kairouz2021advances}. 

\subsubsection{Federated learning with non-IID data}
The canonical FL suffers under non-IID (independent and identically distributed) data distribution. For example, \cite{zhao2018federated} shows that when the client’s data are highly skewed, the accuracy of federated averaging (FedAvg) reduces significantly, by up to $55\%$. Similarly, \cite{li2020federated} points out that FedAvg will not converge to an optimal solution when training data are imbalanced across clients. To make the global model perform better on local data, some FL method adds a few fine-tuning steps to clients' local dataset after federation is completed~\citep{wang2019federated, mansour2020three}. Another stream focuses on modifying the local objective function, and some representative algorithms in this family include FedProx~\citep{sahu2018convergence}, SCAFFOLD~\citep{karimireddy2020scaffold} and FedAMP~\citep{huang2021personalized}. Other actively researched techniques that promote personalized models include setting base layers and personalized layers~\citep{li2021fedbn, arivazhagan2019federated}, meta-learning~\citep{fallah2020personalized}, and knowledge distillation~\citep{li2019fedmd}. These methods mitigate, but cannot eradicate the negative impact caused by data heterogeneity. In this article, we evaluate the impact that data heterogeneity may have on the fairness and incentive design of the FL.

\subsection{Client selection in federated learning}
Client selection refers to a server choosing a subset of qualified clients to participate in an FL partnership~\citep{zeng2021comprehensive}. Existing client selection schemes can be categorized into two types based on their goals. One is to improve communication efficiency, and the other is to improve model effectiveness. The first thread is mostly applied to the cross-device FL, where common practice is to randomly sample a small number of clients to join the federation at the beginning of each round~\citep{nishio2019client, kim2019blockchained, konevcny2016federated}. The convergence properties are usually preserved as this sampling scheme is random, and hence can be considered unbiased. Furthermore, \cite{goetz2019active} proposes a selection mechanism called active FL, where the clients with higher local loss are assigned with a larger probability of being selected for training. A similar idea is also seen in \cite{cho2020client}. Compared to the random sampling scheme, this sampling scheme, although biased, can provide a better convergence rate and even higher accuracy~\citep{cho2020client, goetz2019active}. The second thread is the effectiveness-oriented selection method, which is more relevant to our work, and to choose clients that meet certain performance or computation requirements, commonly seen in the reputation- and auction-based incentive design.
In reputation-based methods, the server uses clients' past behavior to rate the reputation of the clients for the selection criteria~\citep{kang2019toward, liu2011novel}. In auction-based designs, the server broadcasts the bid asks with the selection criteria for participation at the beginning of each round in which the bidders with a high score, rated by their resource qualities and expected payments, are chosen~\citep{zeng2020fmore}. In this paper, we consider a selection strategy based on clients' local performance and choose clients with high local accuracy. By doing so, it can potentially deter free riders from joining the federation and retain high-quality clients (both in terms of data quality and quantity) as well.

\subsection{Incentive design in federated learning}
Early works on FL mostly focus on the optimization and acceleration perspective and made an overoptimistic assumption that all the participants will unconditionally engage in federation without undesirable behaviors such as lack of contribution or withdrawal~\citep{li2020federated, liu2020accelerating}. In practice, however, the contribution choices of participants are driven by many factors such as network communication cost, GPU computation limits, final revenue from FL, and fairness. For example, clients who have high expenses, including the cost of communication and computation, but low rewards would not likely share all the data to avoid a deficit. On the contrary, clients with high revenue but low costs will choose to remain in the loop and keep their strategy unchanged.

Recently, many works have been proposed to design incentive mechanisms to ensure a sustainable and fair federation.
\cite{zhang2022enabling} studies clients' long-term selfish participation behavior and designs an incentive mechanism to reduce the free riders while maximizing the total training data.
\cite{tang2021incentive} proposes an incentive framework to maximize the social welfare which is defined as the total gain of the federation and uses monetary transfer to balance clients' deficit. 
\cite{kang2019incentive} proposes a contract theory-based incentive mechanism for mobile networks to attract high-quality participants. 
\cite{zhan2020learning} studies a Stackelberg game and uses deep reinforcement learning (DRL) to adaptively learn the best policy that maps the participation history (state) to an action that maximizes the expected discounted accumulated reward without knowledge of their decision and accurate contribution evaluation.

In this paper, we proposed a general framework that can be adapted to a variety of cross-silo FL algorithms to incentivize the cooperation between clients. We list the most related works in \cref{tab:related_work} and compare the functionality of each framework. We select these criteria from the following aspect: 1) privacy. ISI is to ensure that minimum private information is required to operate the system. 2) incentive. Fairness, IR, and BB are important considerations when clients make their contribution choice. 3) Performance. DH is a crucial consideration when dealing with non-iid data, and NS can help when the heterogeneity level is beyond the capacity the system can handle. 
Our proposed method, therefore, is designed based on these principles.

\begin{table}[!htbp]
  \centering
  \caption{Comparison of related works with proposed approach}
  \label{tab:related_work}
  \begin{tabular}{l|lllllll}
  \hline
  
  &      
  ISI$^1$ &
  NS$^2$  &  
  Fairness & 
  IR$^3$   &
  BB$^4$   &
  DH$^5$ &\\
  
  \cite{zhang2022enabling} & \tobedone & \tobedone & \tobedone & \done & \tobedone & \tobedone \\
  \cite{tang2021incentive} & \tobedone & \tobedone & \tobedone & \done & \done & \tobedone \\
  \cite{kang2019incentive} & \done & \done & \tobedone & \done & \done & \tobedone \\
  \cite{zhan2020learning} & \done & \tobedone & \tobedone & \tobedone & \tobedone & \tobedone\\
  \textbf{Ours} & \done & \done & \done & \done & \done & \done \\
  \hline
  \end{tabular}
  
  \footnotesize{$^1$Information strongly incomplete: server only have the knowledge of
  the probability that a participant belongs to a certain type but without knowing the private information of users (e.g., costs)
  \\
  $^2$ Node selection
  \\
  $^3$Individual rationality: all the  participants  have  non-negative  profits
  \\
  $^4$Budget balance: the sum of payment for participants is no more than the given budget}\\
  $^5$Data Heterogeneity: client data can be non-IID distributed
  \end{table}
\begin{table}[!htbp]
    \centering
    \caption{Major abbreviations}
    \label{tab:abbr}
    \begin{tabular}{l|l}
    \hline
      Abbreviation       & Description   \\ \hline
    FL   & Federated Learning\\
    ISI   & Information strongly incomplete\\
    NS      &  Node selection\\
    IR     &  Individual rationality\\
    BB     &  Budget balance\\
    \hline
    \end{tabular}
    \end{table}

\section{Background}
\subsection{Federated learning algorithms at a glance}
FL represents a family of algorithms trained in a distributed and collaborative paradigm. In this paper, we mainly discuss the case of the FedAvg algorithm which is widely used and deployed in practice. Interested in the non-IID data setting, we also study a variant called FedBN~\citep{li2021fedbn} as a comparison. We want to note that our proposed framework of client selection and incentive design is general, and thus can accommodate other types of FL algorithms as well.

Consider a scenario of $|\mathcal{N}|$ clients $\mathcal{N} = \{1, ..., N\}$ with distributed datasets $\mathcal{D} = \{D_1, ..., D_{N}\}$, respectively. As shown in \cref{eq:fedavg}, FedAvg aims to minimize the empirical loss over all the data in $\mathcal{D}$ from $\mathcal{N}$, where $n$ is the number of total data points, $n_k$ is the number of data points in client $k$ and $\mb w$ denotes the weights of the model being learned. The overall optimization problem is shown as in \cref{eq:fedavg}, and the pseudocode to solve it is illustrated in \cref{alg:fedavg}. It is of note that the aggregation round defines a period that clients send local models to the global server for model aggregation which is different from the data sharing round we defined in \cref{sec:alg_design}.
\begin{equation}\label{eq:fedavg}
  \min _{\mb w} \sum_{k=1}^{|\mathcal{N}|} \frac{n_k}{n} F_{k}\left(\mb w\right) \quad \text{where} \quad F_{k} = \frac{1}{n_k}\sum_{i=1}^{n_k} \mathcal{L}_{\mb w}(x_i, y_i)
  \end{equation}
  \begin{algorithm}[!htbp]
    \caption{Federated Averging algorithm  (\textcolor{black}{FedAvg} $\|$ \textcolor{red}{FedBN})}
    \label{alg:fedavg}
        \begin{algorithmic}[1]
        \Procedure{FedAvg}{$D_1, ..., D_{N}$}
            \State{Initialize $p_i$ as $\mathcal{D}_i/\sum_{j=1}^{N}D_j$}\Comment{weights of $i^{th}$ client}
            \For{each aggregation iteration $l = 1, 2 \dots L$}
                \State{central server dispatches model $M$ to clients}
                \For{each client $i = 1, 2 \dots N$}
                    \State{$M_i \gets LOCALUPDATE(M, \mathcal{D}_i)$}
                \EndFor
                
                \For{each layer l in $M$ \textcolor{red}{except for batch normalization layer}}
                    \State{$M(l) \gets \sum_{i=1}^{N} p_i M_i(l)$} \Comment{model aggregation}
                \EndFor
                
            \EndFor
        \EndProcedure
        \Statex
        \Statex
        \Function{localupdate}{$M, \mathcal{D}_i$}
            \For{each $\text{epoch}=1, 2 \dots K$}
                \State{Randomly shuffle $D_i$ and create B batches}
                \For{each mini batch $b=1, 2 \dots B$}
                    \State{$w_{b+1} \gets w_{b} - \eta \nabla L_{w_{b}}(X_b, Y_b)$}\Comment{mini batch gradient descent}
                \EndFor
            \EndFor
        \EndFunction
    \end{algorithmic}
\end{algorithm}

\section{Model Formulation}
Consider a cross-silo FL model with $N$ clients. In this section, we describe a client selection procedure after each round of FL. In addition to client selection, we address the issue of fairness in terms of reward distribution among the clients for fairness consideration. \cite{tang2021incentive} proposed a money transfer scheme amongst the clients to build an incentive mechanism to deal with heterogeneity in clients due to varying computational resources. In our formulation, we introduce a similar money transfer mechanism to ensure fair distribution of rewards amongst the participants from the FL model. The proposed formulation simultaneously formulates the client selection and money transfer decisions by solving a constrained optimization problem.

We assume that after every round of FL, each client can receive a ``reward'' using the globally aggregated federated model. These rewards may be explicit after each round, for example, a healthcare organization employing the updated FL model for diagnosis and generating actual dollar revenue. However, in many practical scenarios where FL is employed, the rewards may be realized only after the FL model reaches a certain performance threshold. In such situations, the ``reward" for each client after a data sharing round may be implicit. In such situations, these potential ``rewards" may be used for book-keeping how the explicit rewards must be distributed once they are realized. These potential rewards may be used with an FL incentive mechanism in \cite{yu2020sustainable} that also accounts for the inequality and waiting time for receiving payoffs.

Now, we define the following quantities to formulate the client selection problem:
\begin{enumerate}
    \item \textbf{utility$(n,t)$}: Valuation of the FL model for client $n$ at round $t$. We use the utility measured based on difference of precision proposed by \cite{tang2021incentive}:
    \begin{align}\label{eq:utility}
        \text{utility}(n,t) = u_n\paren{\epsilon(n,t) - \epsilon(n,t-1)}
    \end{align}
    where $\epsilon(t)$ can be a generic precision measure at round $t$ and $u_n$ is revenue per unit increase in precision. In our numerical experiments for a classification task, we take the precision measure to be the accuracy of the global model measured by the clients on local validation sets.
    
    \item \textbf{cost$(n,t)$}: Cost incurred by client $n$ for collecting the data, training the local model, and communicating with the central server at round $t$. For simplicity, we define it as
    \begin{align}\label{eq:cost}
        \text{cost}(n,t) = \underbrace{c_n^{\text{data}}\times s(n,t)}_{(1)} + \underbrace{c_n^{\text{train}}\times K(n,t)}_{(2)} + \underbrace{c_n^{\text{comm}}}_{(3)}
    \end{align}
    (1) represents the cost of data collection, where $s(n,t)$ denotes the number of additional samples collected by client $n$ in between the data sharing rounds $t-1$ and $t$, and $c_n^{\text{data}}$ denotes the unit cost of data collection for client $n$. (2) represents the cost of local model training, where $K(n,t)$ denotes the number of local iterations run by client $n$ at round $t$ and $c_n^{\text{train}}$ denotes the cost of running one iteration of local model update for client $n$. (3) represents the cost of communication with the central server, where $c_n^{\text{comm}}$ denotes the cost of model communication for client $n$.
    
    \item \textbf{mt$(n,t)$}: Money transferred to client $n$ at the end of round $t$. The central server decides this term to control the fair reward distribution amongst the client. We ensure that the money transfer transactions happen in between the clients, with no external input/output of funds, i.e. the budget balance condition is satisfied:
    \begin{align}
        \sum_{n=1}^N \text{mt}(n,t) = 0
    \end{align}
\end{enumerate}

Let $\mathcal{N}=\{1,\dots,N\}$ denote the set of all clients at the start of the FL process. The client selection formulation needs to decide the set of clients to retain $\mathcal{A}(t)\subseteq \mathcal{N}$ after each round $t$ of FL. Once a client is deselected out of the federation, it will not be considered for selection in all future rounds, i.e.,
    \begin{align}\label{eq:leftout}
        \mathcal{A}(t)\subseteq\mathcal{A}(t-1),\ \forall t \in \{1, \ldots, T \}
    \end{align}
    Moreover, the clients deselected out of the federation in a round are not included in the money transfer process during that round, i.e.
    \begin{align}
        \text{mt}(n,t) = 0,\ \forall n\notin\mathcal{A}(t)
   \end{align}

\subsection{Money transfer scheme}
In FL, the clients cooperate to train a global model that is aggregated using local models trained for each individual client. The rewards obtained by using these federated models are realized by each client individually when they deploy these models and generate revenue as a result of that. However, the distribution of these rewards may vary across clients for a variety of reasons such as the heterogeneous nature of data, differing quantities of data contributed, and differing mechanisms for calculating utility and cost. Hence, as discussed in the introduction, there is a need for the federation to set up a fair reward distribution scheme.

While there is a disparity in the reward distribution, the net ``contribution" of each client in training the federated model may also vary. The quantity and quality of the data are two main factors that govern the contribution of each individual model in the FL setting. If we can quantify the contribution of each client, that can form a basis for the reward distribution scheme amongst the clients. We mention below some popular approaches from the literature to quantify the contribution $q(n,t)$ for client $n$ at round $t$:
\begin{enumerate}
    \item \textbf{Quantitative Contribution}: $q(n,t)=\#$ of data points contributed by client $n$ at round $t$. Quantitative contribution is popular due to ease of calculation, but completely ignores the quality aspect~\citep{wang2019measure}.
    \item \textbf{Marginal Contribution}: $q(n,t)=v(\mathcal{N}) - v(\mathcal{N}\setminus\{n\})$, where $v(A)$ measures the utility of the collective $A$. Marginal contribution~\citep{gollapudi2017profit} implicitly takes into account the quality of the data provided by each client in addition to the quantity of data. However, this approach requires evaluating the utility of $N+1$ possible models after each round of FL.
    \item \textbf{Shapley Value}: \\
    $q(n,t)=\sum_{S\subseteq \mathcal{N}\setminus\{n\}} \frac{|S|!(N-|S|-1)!}{N!} \left(v(\mathcal{S}\cup\{n\}) - v(S)\right)$, where $v(A)$ measures the utility of the collective $A$. Shapley value~\citep{wang2019measure, song2019profit} takes into account the marginal contribution of client based on every possible subset of active clients. However, this approach requires a significant computational overhead, since it requires calculation of utility for $2^N$ possible models~\citep{ghorbani2019data}. In practice, when the number of clients $N$ is too large, approximation methods may be considered~\citep{fatima2008linear,liu2022gtg}.
\end{enumerate}

Now that we are able to quantify the contribution from each client, we define a money transfer scheme. The net budget available to the federation for distribution amongst the clients, i.e. the net profit/loss made by the clients, is defined as follows:
\begin{align}
    B(t) = \sum_{n\in\mathcal{A}(t)} \text{utility}(n,t) - \text{cost}(n,t)
\end{align}

To achieve fairness in reward distribution, a money transfer scheme needs to ensure that the pay-off to each client in the active set is proportional to their contribution in the current round, i.e.
\begin{align}\label{eq:reward_fairness}
    \text{payoff}^*(n,t) = \frac{q(n,t)}{\sum_{n\in\mathcal{A}(t)}q(n,t)} B(t),\ \forall n\in\mathcal{A}(t)
\end{align}

The money transfer scheme is defined as follows:
\begin{align}\label{eq:mt}
    \text{mt}^*(n,t) = \text{payoff}^*(n,t) - \paren{\text{utility}(n,t) - \text{cost}(n,t)}
\end{align}

Note that since 
\begin{align}
    \sum_{n\in\mathcal{A}(t)} \text{payoff}^*(n,t) = B(t),
\end{align}
Hence, the budget balance equation
\begin{align}
    \sum_{n\in\mathcal{N}} \text{mt}^*(n,t) = \sum_{n\in\mathcal{A}(t)} \text{mt}^*(n,t) = 0
\end{align}

is satisfied.

\subsection{Client selection formulation}
In this section, we provide a mathematical formulation for client selection and money transfer decisions. We start by introducing the terms Social Welfare and Fairness Consideration that we shall use to define the objective function of the optimization problem.

\subsubsection{Social welfare}
From the perspective of social welfare, the federation wants to maximize the net profit made by the clients. Recall that the net profit made by the clients in the federation, i.e. the total budget of the federation at round $t$ is given by
\begin{align}
    B(t) = \sum_{n\in\mathcal{A}(t)} \text{utility}(n,t) - \text{cost}(n,t)
\end{align}
Therefore, from the social welfare perspective, the federation should select the set of active clients $\mathcal{A}(t)$ such that the budget $B(t)$ is maximized. We can observe that the set of active clients that maximize the budget is given as follows:
\begin{align}
    \mathcal{A}^*(t) = \{n\in\mathcal{A}(t-1) | \text{utility}(n,t) \geq \text{cost}(n,t) \}
\end{align}
i.e. choose the clients who are making a profit at round $t$ and discard the rest of the clients.

\subsubsection{Client selection fairness}
While the objective of social welfare maximization is to obtain the largest possible budget for the federation, it may be conflicting with the client selection fairness consideration, i.e. the federation needs to try and select clients contributing good quality data. The high contributing clients may be making a temporary loss due to a variety of different reasons such as: large cost for collecting data and/or training the local model, relative size compared to the other clients from the federation, etc. As shown above, maximizing social welfare shall eliminate all the clients making a loss in any particular round, including the large contributors. Hence, taking a myopic view of social welfare may hurt the federation in the long term. The contribution aspect can be quantified by looking at the relative contribution of the client to be eliminated compared to that of the retained clients, defined as follows:
\begin{align}
    \frac{q(n,t)}{\sum_{n\in\mathcal{A}(t)}q(n,t)},\ \forall n\in\mathcal{A}(t-1)\setminus\mathcal{A}(t)
\end{align}

These terms can act as ``regularizers" for the social welfare maximization objective to ensure the formulation is less aggressive in eliminating the clients with relatively large contribution. We call the summation of these regularizer terms as client selection fairness objective, defined as follows:
\begin{align}\label{eq:select_fairness}
    \frac{\sum_{n\in\mathcal{A}(t-1)\setminus\mathcal{A}(t)}q(n,t)}{\sum_{n\in\mathcal{A}(t)}q(n,t)}
\end{align}

Note that the client selection fairness consideration is different from fairness in terms of distribution of reward. While the objective function in \cref{eq:select_fairness} governs the client selection fairness, the fairness with respect to reward distribution is enforced by ensuring the payoff to the clients is proportional to their contribution by the constraint in \cref{eq:reward_fairness}.

\subsection{Overall formulation}
Combining the money transfer scheme with the social welfare and contribution fairness terms, we formulate the following optimization problem to obtain the client selection and money transfer scheme:

\begin{align}
    \begin{split}
    \max_{\text{mt}(\cdot,t),\mathcal{A}(t)} & \sum_{n\in\mathcal{A}(t)} \left(\text{utility}(n,t) - \text{cost}(n,t)  \vphantom{\frac{\sum_{n\in\mathcal{A}(t-1)\setminus\mathcal{A}(t)} q(n,t)}{\sum_{n\in\mathcal{A}(t)} q(n,t)}}\right. \\
    & \qquad\qquad \left. - \mu \frac{\sum_{n\in\mathcal{A}(t-1)\setminus\mathcal{A}(t)} q(n,t)}{\sum_{n\in\mathcal{A}(t)} q(n,t)}\label{eq:obj}\right)
    \end{split}\\
    \text{s.t. } & \text{mt}(n,t) = 
    \begin{cases} 
      \text{mt}^*(n,t) & n\in \mathcal{A}(t) \\
      0 & n\notin \mathcal{A}(t)
   \end{cases}\\
   & \mathcal{A}(t)\subseteq \mathcal{A}(t-1)
\end{align}
where $\mu$ is a constant that controls the trade-off between social welfare and contribution fairness terms.

Note that solving the optimization problem for the variables $\text{mt}(\cdot,t)$ is straightforward by the constraint \cref{eq:mt}. Hence, we essentially need to solve a discrete optimization in the variable $\mathcal{A}(t)$. Solving this discrete problem using brute force requires $\mathcal{O}(2^N)$ evaluations of the objective function. The brute-force approach may prove to be computationally challenging for a very large set of clients. However, when the number of clients is not too large, as in a typical cross-silo FL, this approach is feasible. Moreover, the set of clients that can be potentially eliminated, and hence the size of the discrete set for the optimization problem, may be shortened as we shall discuss in the next section.

\subsection{Algorithm design}
\label{sec:alg_design}
In this section, we provide an algorithm for client selection and the money transfer scheme.

\begin{algorithm}[H]
    \caption{Client Selection and Money Transfer Algorithm}
    \label{alg:client_sel}
        \begin{algorithmic}[1]
            \State{Initialize the model weights for the local models}
            \State{Set $\mathcal{A}^*(0)=\mathcal{N}$}
            \For{each data sharing round t=1,\dots,T}
                \State{Run FL model with the set of active clients $\mathcal{A}^*(t-1)$ using \cref{alg:fedavg}}
                \State{Calculate utility, cost and contribution terms for clients in $\mathcal{A}(t-1)$}
                \State{Find the clients that can be potentially eliminated:
         	    $$
         	        \mathcal{E}(t) = \{ n\in\mathcal{A}(t-1) | \text{utility}(n,t)<\text{cost}(n,t)\}\;
         	    $$}
     	        \State{Set $f_\text{max}=-\infty$}
                \For{each $E$ in $\mathcal{P}\paren{\mathcal{E}(t)}$}
                    \State{Calculate the objective $f$ in \cref{eq:obj} for $\mathcal{A}(t) = \mathcal{A}(t-1)\setminus E$}
                    \If{$f \geq f_\text{max}$}
                        \State{Set $\mathcal{A}^*(t) = \mathcal{A}(t-1)\setminus E$}
      		            \State{Set $f_\text{max}=f$}
                    \EndIf
                \EndFor
                \State{Set $\text{mt}^*(n,t)$ for each client according to \cref{eq:mt}}
                
            \EndFor
    \end{algorithmic}
\end{algorithm}

Note that by the property in \cref{eq:leftout}, the set of clients that can be potentially eliminated keeps reducing with the progression of rounds in federated learning. Moreover, note that eliminating a client $n\in\mathcal{A}(t-1)$ who is making a profit at round $t$, i.e. satisfying $\text{utility}(n,t) \geq \text{cost}(n,t)$, hurts the social welfare term as well as the client selection welfare term in the optimization objective given by \cref{eq:obj}. Therefore, the set of clients that can be potentially eliminated at round $t$ can be reduced to the set of clients making a loss at round $t$. This property helps significantly reduce the search space for the discrete optimization problem in the previous section.

\subsection{Evaluation metrics}
The proposed formulation is designed to make the client selection and money transfer decisions based on maximizing a combination of the social welfare term and the selection fairness term. It can be observed that smaller values of $\mu$ lead to larger social welfare but smaller selection fairness, and vice-versa. Hence, the trade-off between the social welfare and selection fairness term is controlled by the parameter $\mu$. In order to demonstrate this trade-off over the length of the federation, we define the following aggregate quantities for measuring social welfare and selection fairness:
\begin{itemize}
    \item \textbf{Total Social Welfare}: Since the social welfare term in the objective function \cref{eq:obj} is cumulative, we define total social welfare as the cumulative sum of the social welfare term, measured up to the current round of federated learning:
    \begin{align}\label{eq:TSW}
        \text{TSW}(t') = \sum_{t=1}^{t'}\sum_{n\in\mathcal{A}(t)} \text{utility}(n,t) - \text{cost}(n,t)
    \end{align}
    \item \textbf{Total Selection Fairness Index}: The selection fairness term in the objective function \cref{eq:obj}, unlike the social welfare term, is not cumulative. We propose to aggregate the selection fairness over the past rounds according to the ratio of the sum of contribution of the selected clients in each round to the overall contribution of all clients:
    \begin{align}\label{eq:TSFI}
        \text{TSFI}(t') =  \frac{\sum_{t=0}^{t'} \sum_{n\in\mathcal{A}(t)}  q(n,t)}{\sum_{t=0}^{t'} \sum_{n\in\mathcal{N}}  q(n,t)}
    \end{align}
\end{itemize}
Now, we will present a toy example to demonstrate the calculations of the evaluation metrics described above. Consider the following simple example:
\begin{center}
\begin{tabular}{ |c|c|c|c|c|c| } 
 \hline
 Round & Client & utility & cost & u - c & q\\
 \hline
 \multirow{3}{*}{Round 1}
     & C1 & 0.2 & 0.1 & 0.1 & 0.4\\ \cline{2-6}
     & C2 & 0.15 & 0.1 &0.05 & 0.2\\ \cline{2-6}
     & C3 & 0.3 & 0.05 &0.25 & 0.4\\ \hline

 \multirow{3}{*}{Round 2}
     & C1 & 0.1 & 0.15 & - 0.05 & 0.5\\ \cline{2-6}
     & C2 & 0.1 & 0.15 & -0.05 & 0.1\\ \cline{2-6}
     & C3 & 0.3 & 0.15 & 0.15 & 0.4\\ \hline
\end{tabular}
\end{center}

Note that in round 1 all the clients satisfy $\text{utility}(n,t) \geq \text{cost}(n,t)$, hence no client is eliminated at the end of round 1. At the end of round 2, clients 1 and 2 do not satisfy $\text{utility}(n,t) \geq \text{cost}(n,t)$. Hence, there are four possible scenarios:
\begin{enumerate}
    \item \textbf{Only Client 1 is eliminated}: The objective function is calculated as follows:
    \begin{align*}
   \sum_{n\in\mathcal{A}(2)} \text{utility}(n,2) - \text{cost}(n,2)  - \mu \frac{\sum_{n\in\mathcal{A}(1)\setminus\mathcal{A}(2)} q(n,t)}{\sum_{n\in\mathcal{A}(2)} q(n,2)} \\
   = 0.1 - \mu
\end{align*}
    We can calculate $\text{TSW} = 0.5$ and $\text{TSFI} = 0.55$
    
    \item \textbf{Only Client 2 is eliminated}: The objective function is calculated as follows:
    \begin{align*}
   \sum_{n\in\mathcal{A}(2)} \text{utility}(n,2) - \text{cost}(n,2)  - \mu \frac{\sum_{n\in\mathcal{A}(1)\setminus\mathcal{A}(2)} q(n,t)}{\sum_{n\in\mathcal{A}(2)} q(n,2)}\\
   = 0.1 - \frac{1}{9}\mu
\end{align*}
    We can calculate $\text{TSW} = 0.5$ and $\text{TSFI} = 0.85$
    
    \item \textbf{Both Client 1 and 2 are eliminated}: The objective function is calculated as follows:
    \begin{align*}
   \sum_{n\in\mathcal{A}(2)} \text{utility}(n,2) - \text{cost}(n,2)  - \mu \frac{\sum_{n\in\mathcal{A}(1)\setminus\mathcal{A}(2)} q(n,t)}{\sum_{n\in\mathcal{A}(2)} q(n,2)}\\
   = 0.15 - \frac{3}{2}\mu
\end{align*}
    We can calculate $\text{TSW} = 0.55$ and $\text{TSFI} = 0.4$
    
    \item \textbf{None of Client 1 and 2 are eliminated}: The objective function is calculated as follows:
    \begin{align*}
   \sum_{n\in\mathcal{A}(2)} \text{utility}(n,2) - \text{cost}(n,2)  - \mu \frac{\sum_{n\in\mathcal{A}(1)\setminus\mathcal{A}(2)} q(n,t)}{\sum_{n\in\mathcal{A}(2)} q(n,2)}\\
   = 0.05
\end{align*}
    We can calculate $\text{TSW} = 0.45$ and $\text{TSFI} = 1$
\end{enumerate}
Hence, eliminating both clients 1 and 2 leads to the largest TSW, but also to the smallest TSFI. Similarly, retaining both clients 1 and 2 leads to the smallest TSW, but also the largest TSFI. Hence, we can observe that the parameter $\mu$ controls the trade-off between social welfare and fairness. Also note that according to the client selection algorithm, the following decisions will be taken based on different values of $\mu$:
\begin{align}
\mathcal{A}^*(2)=
    \begin{cases}
        \{ C3 \} & \text{if } \mu < 0.033 \\
        \{ C1, C3 \} & \text{if } 0.033 \leq \mu < 0.45 \\
        \{ C1, C2, C3 \} & \text{if } \mu > 0.45
    \end{cases}
\end{align}
Hence, setting a larger value for $\mu$ makes the client selection decision more lenient and vice versa.

\section{Experiments}
\subsection{Experimental Setting}
\subsubsection{Datasets}
We conduct the experiments in 3 different settings. The choice of datasets for each setting is described below:
\begin{enumerate}
    \item \textbf{Heterogeneous Data}: We use 5 different digit recognition datasets: MNIST~\citep{lecun1998mnist}, SVHN~\citep{netzer2011reading}, USPS~\citep{hull1994database}, SynthDigits~\citep{ganin2015unsupervised} and MNIST\_M~\citep{ganin2015unsupervised}. Each dataset represents one ``client" for the federated learning task. Sample images from these datasets can be seen in \cref{fig:ex}. While the underlying task for all these datasets/clients is recognizing digits from 0 to 9, the heterogeneity in the form of distributions from which these datasets come from presents an opportunity of simulating a realistic federated learning model.
    \begin{figure*}[!htbp]
    \centering
    \includegraphics[width=\textwidth]{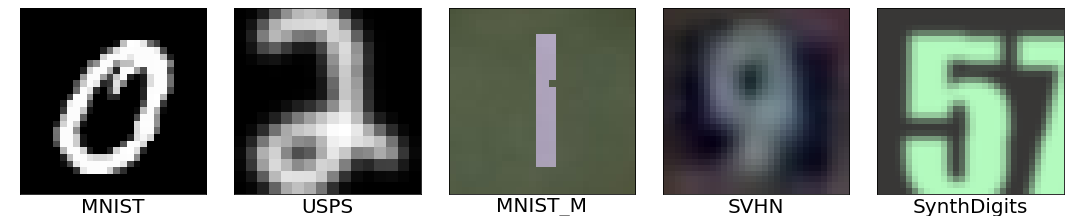}
     \caption{Five different ``client's" data for classification of digits from 0 to 9. MNIST and USPS are grayscale images, while SVHN, SynthDigits, and MNIST\_M are RGB images}
     \label{fig:ex}
    \end{figure*}
    \item \textbf{Homogeneous Data}: We randomly split the MNIST~\citep{lecun1998mnist} and CIFAR-10~\citep{krizhevsky2009learning} datasets into 5 parts each separately. For both datasets, we use each partition as an independent client for the federated learning task. While the data distribution of the images themselves remains roughly the same across clients, we introduce disparity in terms of relative size, i.e. the number of samples contributed by each client. We use this setting to demonstrate the effect of proposed client selection and money transfer formulation on relatively large/small clients in the federation
    \item \textbf{Label Noise}: We randomly split the MNIST~\citep{lecun1998mnist} dataset into 5 parts separately, each partition representing a client. We randomly corrupt the labels for one of the clients with $30\%$ probability to simulate a client with low-quality data and demonstrate the ability of the network to deselect this client out of the federation. 
\end{enumerate}

The number of additional samples to be added for each client during a data sharing round is chosen by Poisson sampling $s(n,t)\sim\text{Poisson}(\lambda)$ to mimic randomness in the sample size found in the real world scenario. For digit recognition tasks, the mean sample size is held constant at $\lambda=100$ across different clients and different data sharing rounds of federated learning. Note that the digit recognition problem is relatively easier in comparison to modern machine learning tasks, and a large sample size can cause federated learning models to saturate very quickly. Hence, the sample size for the digit recognition tasks is kept relatively small in order to demonstrate the effectiveness of the federated learning task. The mean sample size in the homogeneous setting varies for different experiments and is described in detail in the results below. During each data sharing round, $30\%$ of the new data acquired by the clients is randomly split and added to a local validation set, while the rest of $70\%$ is added to the training set. The local update for the federated learning algorithm is performed using the training set, and the performance evaluation metrics such as utility and contribution terms are calculated using the validation set locally.

\subsubsection{Federated averaging model setting}
We use a six-layered convolutional neural network with 3 convolutional and 3 fully connected layers, with BN layers following every feature extraction layer. We use a stability-based early stopping strategy while performing the aggregation iterations in FedAvg/FedBN models. In order to save communication costs, we halt the aggregation iterations when the change in validation accuracy for each client is smaller than $1\%$ after an aggregation iteration. The maximum number of aggregation iterations is chosen to be at most 5. The batch size for the local model update is 32. For the heterogeneous case, FedAvg algorithm is used for the FL task, whereas for the heterogeneous case, we use FedBN Algorithm.

\subsubsection{Client selection and money transfer algorithm setting}
The FL model is designed to run up to a maximum of $15$ data sharing rounds. Note that the FL model may terminate earlier if the client selection formulation eliminates all of the clients (or all but one) from the federation. The utility term for each client is calculated using \cref{eq:utility}, with the constant $u_n=1$, and the precision measure $\epsilon(n,t)$ is chosen to be the federated model accuracy for client $n$ measured on a locally held-out validation set at the data sharing round $t$. The cost term for each client is calculated using \cref{eq:cost}, with the constant $c_n^{\text{data}}=2\times 10^{-4}$. We assume that the clients do not bear the cost of training and communication, i.e. $c_n^{\text{train}}=c_n^{\text{comm}}=0$. The contribution of each client is calculated using marginal contribution strategy: $q(n,t)=v\paren{\mathcal{A}(t-1)} - v\paren{\mathcal{A}(t-1)\setminus\{n\}}$, where collective utility $v(A)$ is measured as the mean validation accuracy for the federated model obtained by aggregating the local models of the clients in the set $A$.

\subsection{Duration of the federation}
\begin{figure}[!h]
    \centering
    \includegraphics[width=0.5\textwidth]{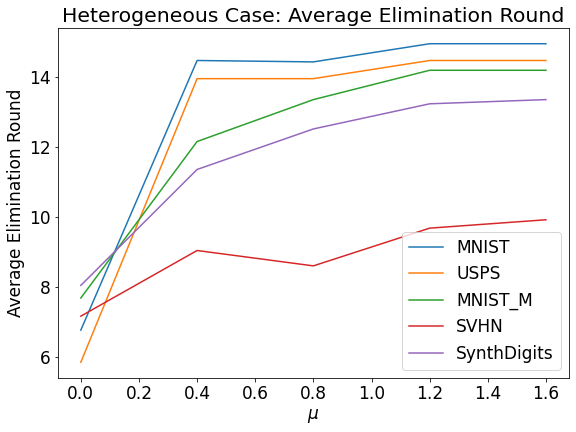}
     \caption{The average elimination round for each client for different values of the parameter $\mu$ in the heterogeneous data setting. The results are averaged over $100$ replications}
     \label{fig:digit_het}
    \end{figure}

The proposed method introduces a money transfer scheme to ensure a fair distribution of rewards under the FL model. The fairness aspect plays the role of incentivizing clients to continue collaborating with others in the federation for a longer duration. We compare the duration for which each client is collaborating in the federation using the proposed method compared to the following two heuristic client selection formulations:
\begin{itemize}
    \item \textbf{Least lenient}: At each round, select the clients making a profit and eliminate the ones making a loss. While this strategy will maximize the social welfare/net profit of the federation at each round, it would also lead to the early elimination of some clients
    \item \textbf{Most lenient}: Never eliminate any client from the federation. This ensures that each client stays in the federation until the end; however, the social welfare of the federation will suffer
\end{itemize}
Note that the first approach can be obtained by setting $\mu=0$ in the proposed formulation, whereas the second approach can be obtained by setting $\mu\rightarrow\infty$. Hence, the parameter $\mu$ controls the leniency of the formulation in terms of eliminating clients from the federation. In \cref{fig:digit_het}, we show the duration spent by each client in the federation for different values of $\mu$ in the heterogeneous data setting. We can observe that setting a larger value for $\mu$ allows a longer duration for clients to stay in the federation.

    \begin{figure*}[!h]
    \centering
    \includegraphics[width=\textwidth]{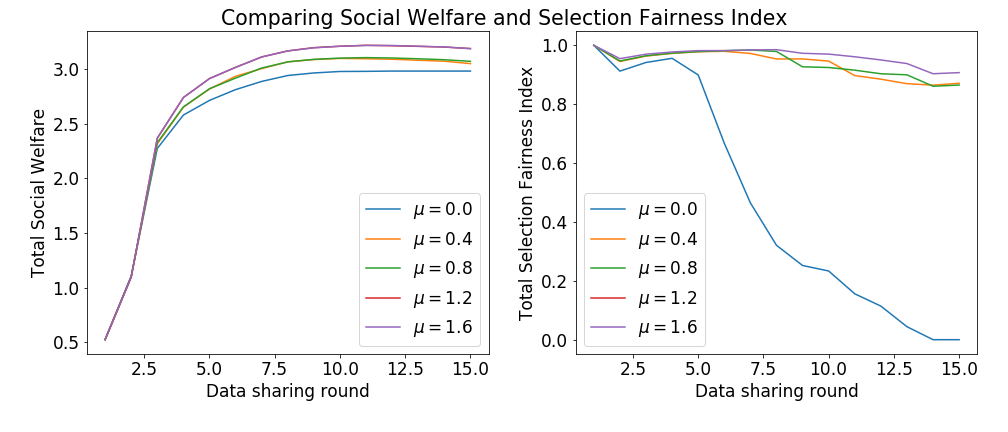}
     \caption{Evolution of Total Social Welfare and Total Selection Fairness Index for different values of $\mu$.  The results are averaged over $100$ replications}
     \label{fig:SW_FI}
    \end{figure*}
     We can observe that TSW rises quickly in the initial rounds of federated learning. However, once the model saturates it may decrease because of clients making losses. Note that setting smaller values for the parameter $\mu$ leads to a higher likelihood of eliminating clients making a loss, resulting in a larger social welfare term in the optimization objective. However, it may hurt the federation in the long term if the eliminated client incurred a loss due to randomness rather than stagnating model performance. We observe in \cref{fig:SW_FI} that in the later data sharing rounds, the total social welfare increases on average by increasing the parameter $\mu$. This demonstrates that the federation may benefit from retaining the clients making a temporary loss. Moreover, smaller values of $\mu$ parameter also lead to decreased total selection fairness index, signifying that clients with a relatively large contribution may be eliminated in order to maximize the social welfare term temporarily.

    \begin{figure*}[!h]
        \centering
        \includegraphics[width=\textwidth]{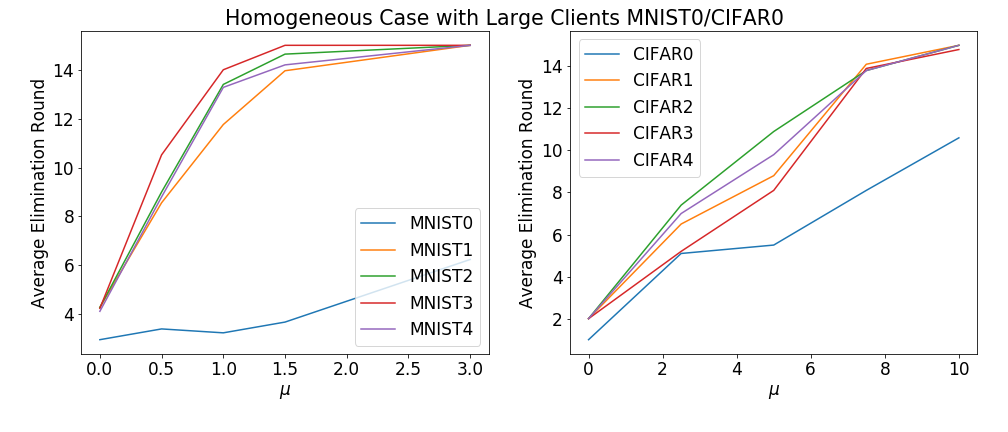}
         \caption{The elimination round for each client for different values of $\mu$ under a homogeneous setting for MNIST and CIFAR-10 data clients. The relative size of clients MNIST0 and CIFAR0 is 5 times that of the rest of the clients. The results are averaged over $100$ replications}
         \label{fig:hom_large}
        \end{figure*}

    \subsection{Social welfare and fairness tradeoff}
    The proposed formulation is designed to make the client selection and money transfer decisions based on maximizing a combination of the social welfare term and the contribution fairness term. We have seen that the parameter $\mu$ plays the role of the leniency parameter: decreasing $\mu$ leads to larger social welfare and vice-versa. However, since small values of $\mu$ lead to earlier elimination of clients, it increases the client selection fairness term, hurting the overall objective. Hence, there is an inherent trade-off between the social welfare and client selection fairness terms in the individual rounds which is controlled by the parameter $\mu$. We have defined the quantities Total Social Welfare(\cref{eq:TSW}) and Total Selection Fairness Index(\cref{eq:TSFI}) to aggregate the social welfare and client selection fairness terms over past rounds of federated learning. In the \cref{fig:SW_FI}, we show the evolution of these metrics across different values of the parameter $\mu$

        \begin{figure*}[!h]
        \centering
        \includegraphics[width=\textwidth]{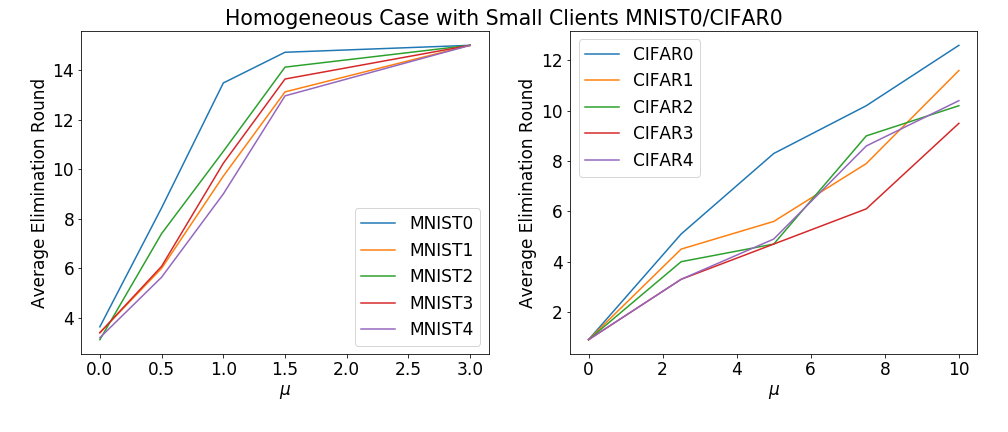}
         \caption{The elimination round for each client for different values of $\mu$ under a homogeneous setting for MNIST and CIFAR-10 data clients. The relative size of clients MNIST0 and CIFAR0 is 0.5 times that of the rest of the clients. The results are averaged over $100$ replications}
         \label{fig:hom_small}
        \end{figure*}
    
     \subsection{Unequal client sizes}
     In this section, we explore the cases where the clients have unequal sizes in terms of the number of samples collected in between each data sharing round. We perform the experiments in the homogeneous data setting, i.e. each client's data is randomly sampled from the same dataset. We consider two scenarios:
     \begin{enumerate}
         \item \textbf{Large Client}: Client 0 contributes on average 5 times the amount of data in comparison to other clients on average. The sample size of Client 0 for each data sharing round $t$ is chosen by Poisson sampling $s(0,t)\sim\text{Poisson}(300)$, while sample size for other clients for each data sharing round is chosen by Poisson sampling $s(n,t)\sim\text{Poisson}(60)$
         \item \textbf{Small Client}: Client 0 contributes on average 0.5 times the amount of data in comparison to other clients on average. The sample size of Client 0 for each data sharing round $t$ is chosen by Poisson sampling $s(0,t)\sim\text{Poisson}(60)$, while sample size for other clients for each data sharing round is chosen by Poisson sampling $s(n,t)\sim\text{Poisson}(120)$
     \end{enumerate}

     We simulate a total of 5 clients with the sampling criteria mentioned above for MNIST and CIFAR-10 datasets. In \cref{fig:hom_large}, we demonstrate the average elimination rounds for each client in large client scenario. Note that the utilities for each client in the homogeneous data case are roughly equal, since the utility for each client is measured as the validation accuracy of the federated learning model, and the validation sets for each client are sampled from the same underlying distribution. However, the costs for client 0 are on average 5 times larger in comparison to other clients. Hence, client 0 is more likely to make a loss and potentially get eliminated. Since digit classification for the MNIST dataset is a relatively easier classification task and requires a fewer number of effective samples to stabilize the federated learning model, we notice that the client MNIST0 gets eliminated much earlier in comparison to other MNIST clients. For the CIFAR-10 dataset, the effective number of effective samples to stabilize the federated learning model is relatively larger. Hence, the client CIFAR0 is not eliminated immediately, but it is eliminated before other CIFAR clients. In \cref{fig:hom_small}, we demonstrate the small client scenario. Contrary to a large client scenario, the costs for client 0 are on average half of that of other clients. Hence client 0 is less likely to incur a loss and potentially get eliminated. In \cref{fig:hom_small} we observe that the average elimination round for client 0 is larger in comparison to other clients, as expected.

        \subsection{Selecting high quality clients}
        A key property of measuring the contribution of each client using the marginal contribution method is to account for both the quantity and quality of data simultaneously. In this section, we demonstrate that the client selection fairness objective in the proposed formulation coupled with contribution measured using the marginal contribution method is able to select high-quality clients. We perform the experiments in the homogeneous data setting for the MNIST dataset with equal client sizes. However, in order to simulate a client with low quality, we randomly corrupt the labels of $30\%$ of the samples for client 0.
        
\begin{figure}[!h]
        \centering
        \includegraphics[width=0.5\textwidth]{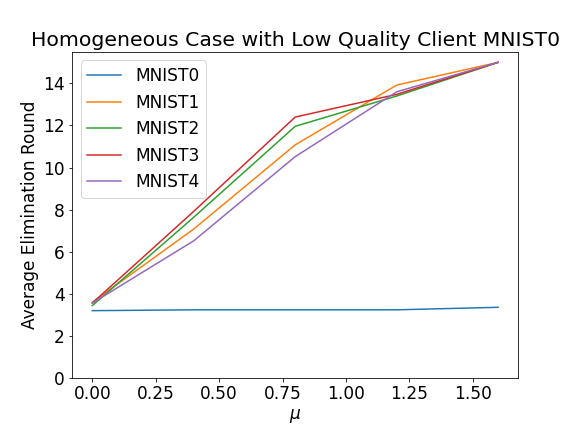}
         \caption{The elimination round for each client for different values of $\mu$ under a corrupted data setting for MNIST data clients. The labels for client MNIST0 are randomly corrupted with $30\%$ probability. The results are averaged over $100$ replications.}
         \label{fig:MNIST_corr}
        \end{figure}

        The contribution of each client is calculated using marginal contribution method, i.e. $q(n,t)=v\paren{\mathcal{A}(t-1)} - v\paren{\mathcal{A}(t-1)\setminus\{n\}}$, where collective utility $v(A)$ is measured as the mean validation accuracy for the federated model obtained by aggregating the local models of the clients in the set $A$. We would expect the collective utility $v\paren{\mathcal{A}(t-1)\setminus\{0\}}$ to be larger than the collective utility $v\paren{\mathcal{A}(t-1)\setminus\{n\}}$ for other clients $n$ since the sample size for each client is roughly equal, but data for client 0 has lower quality due to label noise. Hence, the relative contribution of client 0 is smaller in comparison to other clients. So when client 0 incurs a loss, it is easier for the client selection formulation to eliminate it due to a smaller impact on the client selection fairness term(\cref{eq:select_fairness}) in the objective compared to other clients. In \cref{fig:MNIST_corr}, we show the average elimination rounds for each client. We observe that the low-quality client 0 is eliminated very early, whereas the average elimination round for other clients grows steadily as we grow the parameter $\mu$. 
        
        \section{Discussion}

        In this work, we studied the incentive and fairness issues in cross-silo federated learning. We introduced a client selection scheme to keep clients with high data quality in the loop while removing clients with low quality. To ensure fairness in reward distribution, we also introduced a money transfer scheme in which the pay-off to a client is proportional to its contribution in the current round. The experimental results on a digit recognition task show that our proposed framework can achieve high social welfare (i.e., total utility gain of the federation) by promoting a long-term FL partnership, while maintaining fair distribution of rewards to participants. Interestingly, we also found that the hyperparameter of $\mu$ plays an important role in balancing social welfare and client selection fairness, which can be adjusted to accommodate different real-world situations.
        
        We note that there are some limitations of our work. For example, we only experimented in a simulated federated learning environment which may not perfectly reflect the real-world situation. We assumed the cost of each client is given and the utility is determined by the local model performance only, while in practice the cost may not be easily accessible and the utility can be determined by multiple factors. For our future work, we plan to test the incentive mechanism under more federated learning algorithms in a real-world environment. We are also interested in extending our work to the cross-device setting where scalability and system heterogeneity also need to be considered.

\section*{Acknowledgments}

The research was partially supported by Cisco Systems, Inc and by NSF CMMI 2038403. The authors thank the Co-Editor-in-Chief and the anonymous referee for their insightful comments, which improve the article significantly. 

\bibliography{ref}
\bibliographystyle{imsart-nameyear}


%
%

\end{document}